\pdfoutput=1
\documentclass[letterpaper, 10 pt, conference]{ieeeconf}  

\IEEEoverridecommandlockouts                              
\usepackage[space]{cite}

\usepackage{graphicx} 
\usepackage{xcolor}

\begin{document}
\thispagestyle{empty}
\pagestyle{empty}

\title{\LARGE \bf Toward Robotically Automated Femoral Vascular Access}

\author{Nico Zevallos,$^{*,1}$ Evan Harber,$^{*,1}$ Abhimanyu$^{1}$, Kirtan Patel$^{2}$, Yizhu Gu$^{1}$, Kenny Sladick$^{1}$, Francis Guyette$^{3,5,6}$, \\Leonard Weiss$^{3,5,6}$, Michael R. Pinsky$^{4}$, Hernando Gomez$^{4}$,  John Galeotti$^{1}$, and Howie Choset$^{1,2}$

\thanks{This work was supported by DoD BAA W811XMH-18-SB-AA1 BA180061 and W81XWH-19-C-0101}   

\thanks{$^{*}$Authors of Equal Contribution}

\thanks{$^{1}$Robotics Institute, Carnegie Mellon University, 5000 Forbes Avenue, Pittsburgh, USA {\tt\small nzevallo@andrew.cmu.edu, jgaleotti@cmu.edu, choset@andrew.cmu.edu}}

\thanks{$^{2}$Department of Mechanical Engineering, Carnegie Mellon University, 5000 Forbes Avenue, Pittsburgh, USA}

\thanks{$^{3}$Department of Emergency Medicine, University of Pittsburgh, Pittsburgh, PA, 15261, USA}

\thanks{$^{4}$Department of Critical Care Medicine, University of Pittsburgh, Pittsburgh, PA, 15261, USA}%

\thanks{$^{5}$UPMC Emergency Department Attending Physician. $^{6}$STAT MedEvac}}%







\maketitle



\begin{abstract}

    Advanced resuscitative technologies, such as Extra Corporeal Membrane Oxygenation (ECMO) cannulation or Resuscitative Endovascular Balloon Occlusion of the Aorta (REBOA), are technically difficult even for skilled medical personnel. This paper describes the core technologies that comprise a teleoperated system capable of granting femoral vascular access, which is an important step in both of these procedures and a major roadblock in their wider use in the field. These technologies include a kinematic manipulator, various sensing modalities, and a user interface. In addition, we evaluate our system on a surgical phantom as well as in-vivo porcine experiments. These resulted in, to the best of our knowledge, the first robot-assisted arterial catheterizations; a major step towards our eventual goal of automatic catheter insertion through the Seldinger technique. 

\end{abstract}

\section{Introduction}

    
    
    Robotic surgery is a mature field with years of research \cite{Robotics:Taylor2016, Robotics:Okamura2017, Robotics:Simaan2009, Robotics:Simaan2018, Robotics:Dario2003}, inventions \cite{Robotics:Simaan2013, Robots:DaVinci, Robots:Medrobotics2012} and successful use in medicine \cite{Robotics:Beasley2012, Robots:DaVinci, Robotics:Burgner-Khars2015}. These surgical tools strive to significantly augment the surgeon's perception and control over surgical instruments. A trained surgeon's skills can be greatly enhanced by tools such as tremor reduction, virtual fixtures, or motion scaling. These technologies grant a significant edge over classical surgical techniques, especially in the field of endoscopic surgeries \cite{Robots:DaVinci}. 
    
    These systems have also been extended to automate certain repetitive tasks to reduce the cognitive burden on surgeons \cite{Robots:Zhang2010, Robots:Zevallos2017, Robots:vandenBerg2010, stiffness}. In many of these cases teleoperated systems are adapted to increase the level of automation, moving towards more supervised robotic surgeries. 
    
    \begin{figure}[thpb]
        \centering
        \includegraphics[width=3.2in]{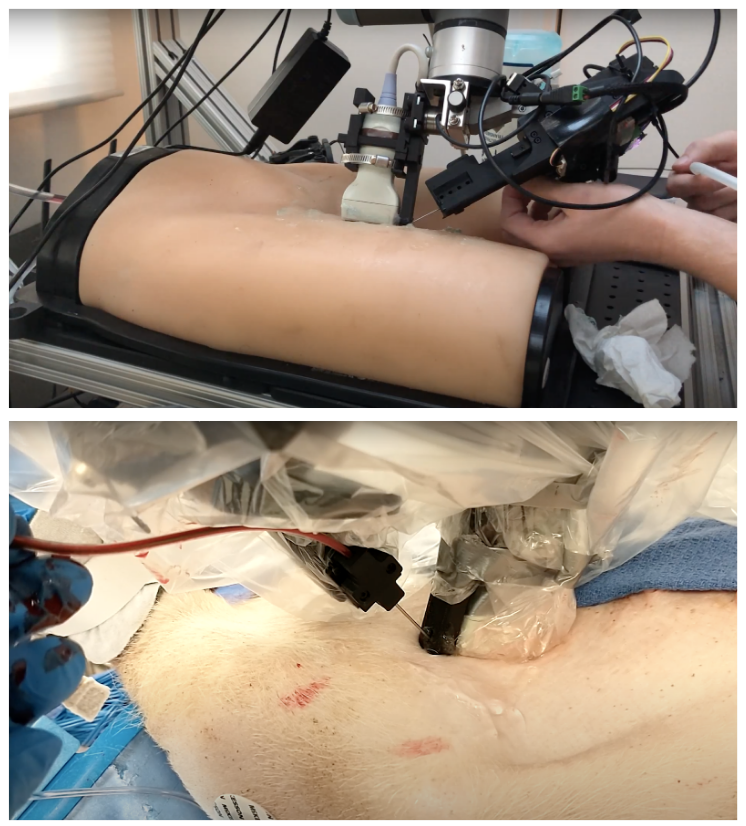}
        \caption{The two degree of freedom (dof) end effector used for femoral access mounted on a 6 dof Universal Robotics UR3e robotic arm. The system was validated by inserting a guide wire into both a pair of leg phantoms (top) as well as a series of porcine (pig) experiments (bottom).}
        \label{fig:photo}
    \end{figure}
    
    Robotic needle insertion is also not a new area of research. By automating control of needles through a combination of tracking and steering methods robotic needle insertion can potentially increase accuracy for precise medical procedures. For instance, more accurate placement of needles during biopsies can lead to fewer false negatives. There have been many different approaches to automating needle insertion, and until recently none have been successfully applied to clinical systems \cite{Robots:Elgezua2013}. In 2020 \cite{Robots:Chen2020} developed a table-top system guided by deep learning to autonomously draw a patient's blood from their arm which was successfully tested on human subjects.
    
    In addition to this research, commercial products have been developed to assist medical professionals in inserting needles. Some use a needle guide to help the user keep a fixed angle with respect to the ultrasonic probe\cite{Seldinger:NationalUltrasound2020}. Others help visualize vein position using near-infrared light similar to technologies used in robotic systems\cite{Robots:Chen2020}. These systems are sensitive to differences in skin color\cite{Seldinger:Woude2013} and limited to veins up to 1 cm below the skin \cite{Seldinger:Venilite2020, Seldinger:AccuVein2020} due to the absorption rate of infrared (IR) light by water in the tissue\cite{Seldinger:Jobsis1977}.
    
    Neither existing minimally-invasive surgery tools nor recently developed vascular access systems are well suited for automated femoral access. Performing the Seldinger technique on the femoral artery presents several challenges unique to the anatomy of the femoral region.  First and foremost is the size and variability of the area scanned. Good femoral access means maneuvering instruments around the highly variable groin anatomy as we search to determine the best (or even multiple) insertion points for a robotic system. A robot could have to travel down the leg or reach around anatomy with sensors and mechanisms attached in a way that small-work-space laparoscopic tools built to operate inside the body are not suited for. For example, the table-top designs of \cite{Robots:Chen2020} would have difficulty conforming to the complex curved anatomy of the groin. 
    
    Another difference is depth. Most minimally invasive systems operate through an incision on the skin or a natural orifice, and while they are technically able to operate outside the body, current ultrasonic sensors built to reach the depths necessary for femoral access are not yet compact enough to fit on the dexterous final joints that afford laparoscopic robots their last degrees of freedom. In \cite{Robots:Chen2020}, the vessels can be detected visually from the surface of the skin, but as mentioned previously, this only work when vessels are superficial. As the femoral artery extends distally from the inguinal ligament (An area about 4 cm long where vascular access is most feasible) it may vary in its course and depth and may be obscured by 2-8 cms of fat and connective tissue \cite{Seldinger:Gopalakrishnan2019}. The combination of varying tissue depth and a range of possible of surgical sites lead us to seek a design (Fig. \ref{fig:photo}) that would be flexible enough to account for the many unknown variables that we could encounter.
    
    Using the system seen in Fig. \ref{fig:photo}, a single operator with no medical experience was able to gain femoral vascular access in-vivo. To the best of our knowledge this marks the first instance of a robotically assisted femoral catherization. In this paper, we will present an overview of the Seldinger technique, the medical technique we are attempting to automate, a technical breakdown of the software and hardware components required for each sub-task, and an overview of our experimental results on pigs as well as surgical phantoms. This system, though not autonomous, is the first step in the development of a fully automated system.

\section{Background}

    \subsection{Vascular Access}

    Obtaining vascular access during the resuscitation of a critically ill patient is challenging especially in austere environments. Current strategies rely on skilled providers to place intravenous (IV) and intraosseous (IO) access. These skills are not universally available. In cases of cardiac arrest or trauma, repeated attempts to secure vascular access may delay transport to definitive care or take the provider away from critical tasks such as CPR or emergent hemorrhage control. In military or disaster response moving additional providers into the disaster scene or the battlespace puts the rescuers at risk. Ideally vascular access could be achieved automatically, freeing the provider to focus on other tasks. 
    
    In addition, advanced resuscitative technologies such as Extra Corporeal Membrane Oxygenation (ECMO) cannulation, a procedure used to oxigenate the blood during possible lung or heart failure, or Resuscitative Endovascular Balloon Occlusion of the Aorta (REBOA), which stops serious hemoraging by inserting a balloon catheter into a large vessel, are technically difficult even for skilled providers. Use of an ultrasound guided robotic system to obtain vascular access and place these catheters will enable these procedures to be performed in areas without highly skilled providers and improve timely access to life saving interventions. In both REBOA and ECMO, the rate limiting step is safe cannulization of the femoral artery \cite{Seldinger:Chonde2020}. This task is typically performed by one or two physicians in the controlled environment of a resuscitation bay or operating room \cite{Seldinger:Bulger2019}. An ultrasonic guided robot can allow these life-saving procedures without a highly skilled operator. Through integrating machine learning, ultrasound signal processing, force sensing and modular robotics can make safe cannulation by a minimally skilled operator possible.
    
    
\subsection{Seldinger Technique}\label{section:seldinger}

    \begin{figure}[thpb]
        \centering
        \includegraphics[width=2.5in]{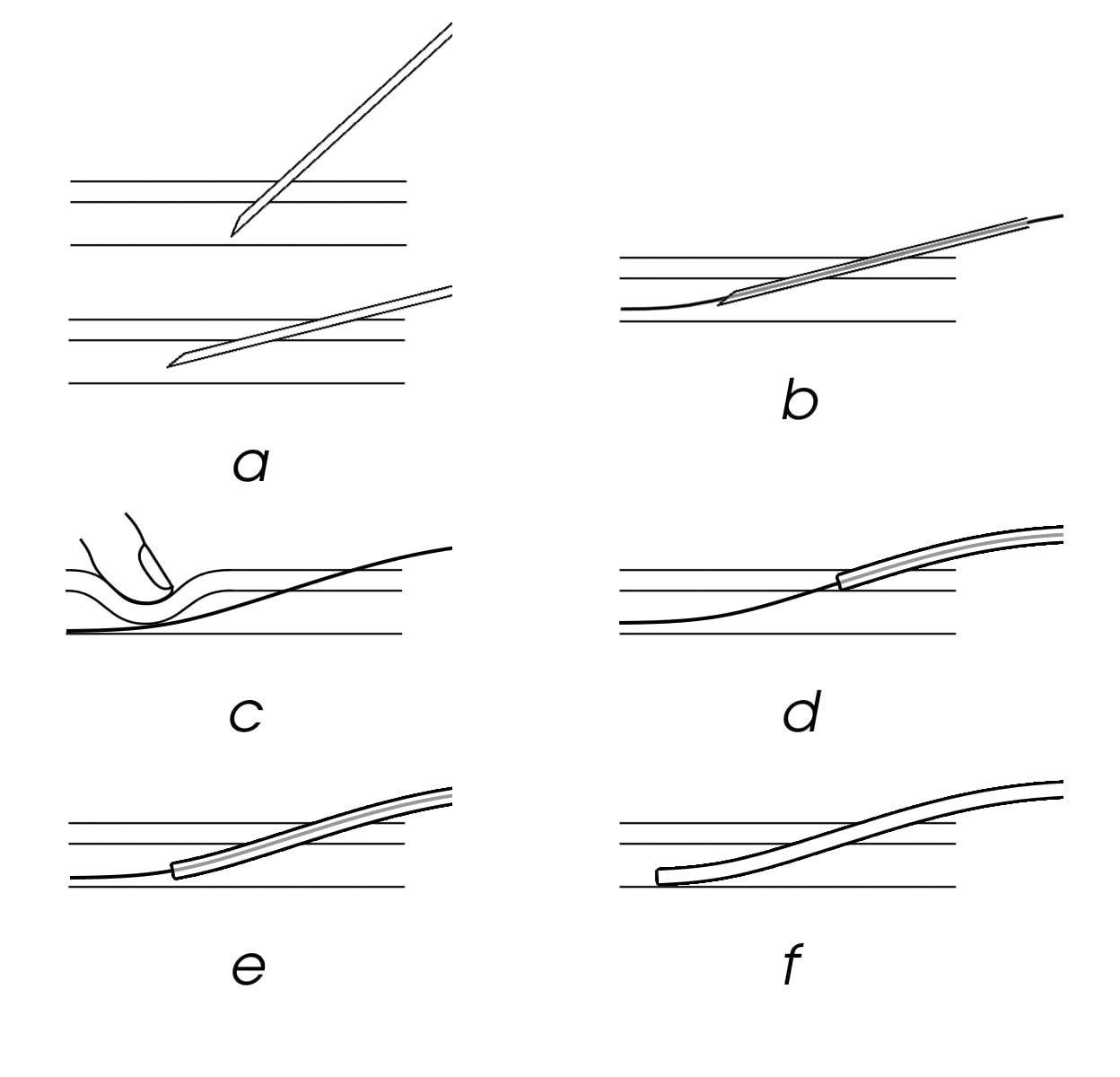}
        \caption{The modified Seldinger technique as originally described in his 1953 paper \cite{Seldinger:Seldinger1953}. a) The vessel is located by ultrasound guidance and punctured. b) A guidewire is inserted through the needle into the vessel beyond the needle tip. c) The needle is withdrawn leaving the guidewire in the vessel. d) A  catheter is threaded over the guidewire into the lumen. e,f) The guidewire is withdrawn.} 
        \label{fig:seldinger}
    \end{figure}

    

    The Seldinger technique \cite{Seldinger:Seldinger1953} (Fig. \ref{fig:seldinger}) is a surgical method for placing a catheter into a vessel. It was developed in the 1950s, replacing previous methods that involved either exposing the artery surgically using a blunt cannula or large-bore needles that could carry a catheter inside them. While previous techniques often inflicted damage to the vessel wall, Seldinger technique avoids these complications by inserting a needle parallel to the skin directly into the vessel  (Fig. \ref{fig:seldinger}a) and tilted to be more parallel to the skin. Then, a flexible guide wire is inserted through the needle (Fig. \ref{fig:seldinger}b). The needle is removed (Fig. \ref{fig:seldinger}c,d) and the catheter is slid over the guide wire into the artery (Fig. \ref{fig:seldinger}e,f). This method has been further augmented with the use of sagital and transverse ultrasound to guide the needle insertion, with better outcomes for the patient\cite{Seldinger:Loon2018, Seldinger:Stolz2015}. This method is not foolproof, however, and can still cause complications when performed by professionals\cite{Seldinger:Morata2020}. Although more advanced techniques such as ECMO and REBOA require serial tissue dilations and complex insertions at the access point, both these techniques or even the most basic IV placement depend on this primary coordinated effort to assure functional success.

\section{Insertion Mechanism}

     Our system for teleoperated needle insertion, as seen in Fig. \ref{fig:photo}, aims to mechanize the Seldinger Technique. It can be broken down into two major components highlighted in Fig. \ref{fig:ee_current}:
    
    \begin{enumerate}
        \item Ultrasound scanning (blue): an ultrasound probe, robot arm, and force sensor for ultrasound based perception.
        \item Needle Insertion (red): a linear and angular actuator for needle insertion.
    \end{enumerate}

\subsection{Hardware}
\begin{figure}[thpb]
        \centering
        \includegraphics[width=3.5in]{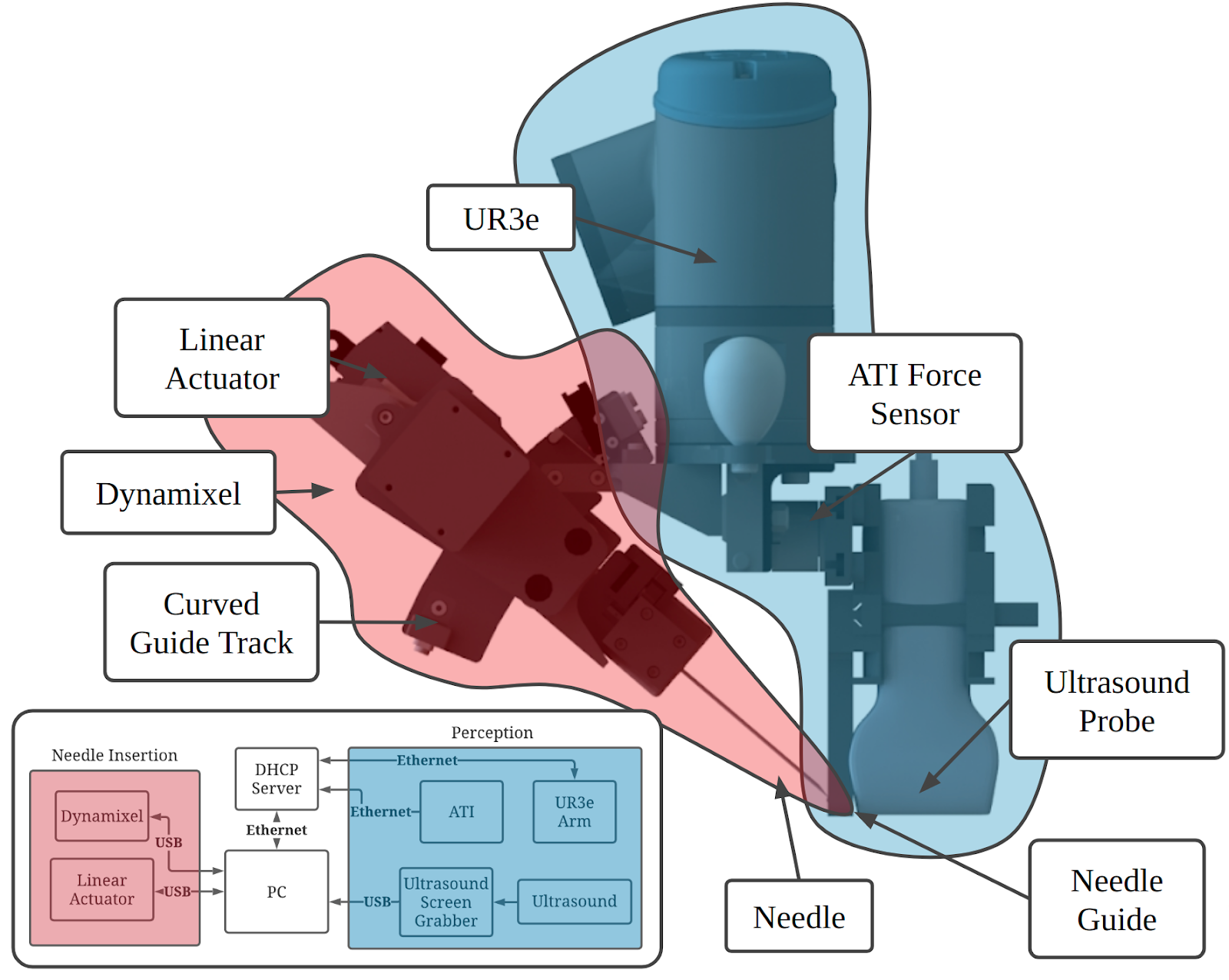}
        \caption{Major components of the end effector. Red specifies the region related to needle insertion, while blue specifies the region associated with ultrasound scanning. All components are connected to a PC which takes in data and coordinates high level commands between the components.}
        \label{fig:ee_current}
    \end{figure}

    In practice femoral access is a multi handed, sometimes multi-physician task. This can pose many issues in the field of robotics where coordination between robotic systems can be very computationally expensive. To reduce the amount of manipulation and robotic degrees of freedom required we designed our hardware around single-manipulator approach. This decision has led us to the design seen in Fig.\ref{fig:ee_current}, where both perception and insertion occur using the same manipulator. 
    
    The major sections of the mechanism can be broken down into its two functionalities: 
    
    \subsubsection{Ultrasound Scanning (blue)} 
    
        To gain subdermal information a Fukuda Denshi portable point-of-care ultrasound scanner was attached to a UR3e arm. Due to the workspace constraints of the UR3e arm the probe was mounted vertically giving the system the greatest amount of manipulatibity through the smallest possible form factor.
    
    
        We incorporated a 6 axis ATI Nano25 force sensor which fits between the ultrasound probe and the UR3e arm. Using isolated force feedback which only measures force along the ultrasound probe leads to better, more accurate ultrasound scanning. 
    
    \subsubsection{Needle Insertion (red)}
    
        The second half of the mechanism relates to insertion control. The main component on this side is the PA-12-22017512R linear actuator and angular degree of freedom. The angular degree of freedom consists of a belt-driven carriage on a curved guide track actuated by a M42P-10-S260-R dynamixel motor. The curve of the guide track was designed to rotate around the point of needle insertion to adjust the needle's angle with respect to the surface of the skin. 
        
        The other major component of the insertion mechanism is the needle guide which helps keep the needle aligned with the plane of the ultrasound image during insertion.

    \subsection{Data streams}

        In total the system contains 3 forms of actuators: (i) the PA-12 Linear actuator used for needle insertion, (ii) a dynamixel motor to control the angular dof which are commanded and give feedback over USB, and (iii) a UR3e robotic arm which is controlled through Ethernet. The ultrasound images are collected through an AV.io HD screen grabber which converts the S-video output of the ultrasound machine to video over USB. The ATI sensor gives force feedback over ethernet. All of these data streams are passed into a PC which converts them to time stamped ROS topics used for central data collection and control. The data diagram for the system can be found in Fig. \ref{fig:ee_current}.
        
\section{Femoral Access Procedure}\label{procedure}

        Along with these physical components, velocity-position and image-based control software were designed to automatically ultrasonically scan over a surface, mark an insertion location, adjust the angle of the probe, and insert a needle.

        The scanning and insertion procedures can be summarized as follows:
        
        \begin{enumerate}
            \item Move robot to select points along scanning trajectory.
            \item Autonomously scan along the desired trajectory while keeping an eye out for a good point for insertion. 
            \item Use the GUI to return to the desired insertion point. 
            \item Center the desired vessel in the ultrasound image using the image space controller while simultaneously rotating until the vessel is transverse in the ultrasound image. 
            \item Use the Needle calibration to set the desired angle then the desired insertion depth to puncture the vessel and insert. 
            \item Fine tune the insertion depth using the needle tweak controller until the needle is located in the center of the vessel.
            \item Manually insert the guide wire through the back of the needle.
            \item Retract the needle and move the robot away from the skin leaving the guidewire inserted in the vessel. 
        \end{enumerate}
        
\subsection{Ultrasound scanning (1-3)}
    
    One of the core components of the system was the ability to perform smooth, even scans of the femoral region. Constant pressure was important to maintain good ultrasonic coupling necessary for imaging. These scans, though not strictly necessary for insertion, help the operator get a situational awareness of the underlying anatomy. This scanning data will also help in the future as we train automatic vessel detection algorithms.
    
    To begin a scan, a trajectory of scanning locations were manually chosen on the surface of the surface based on the choice of the artifact to be captured using ultrasounic imaging. Then starting from a safe position a hybrid force-position controller was used for scanning the surface. The force controller ensured a constant contact of the ultrasound probe with the patient's skin surface. A position controller drove the ultrasound controller from the initial point to the final point of scan while moving compliantly in the vertical axis of the ultrasonic probe. The controller is explained in more detail in \cite{Robots:Ed2020}. 
    

\subsection{Image-space controller (4)}

    \begin{figure}[thpb]
        \centering
        \includegraphics[width=2.2in]{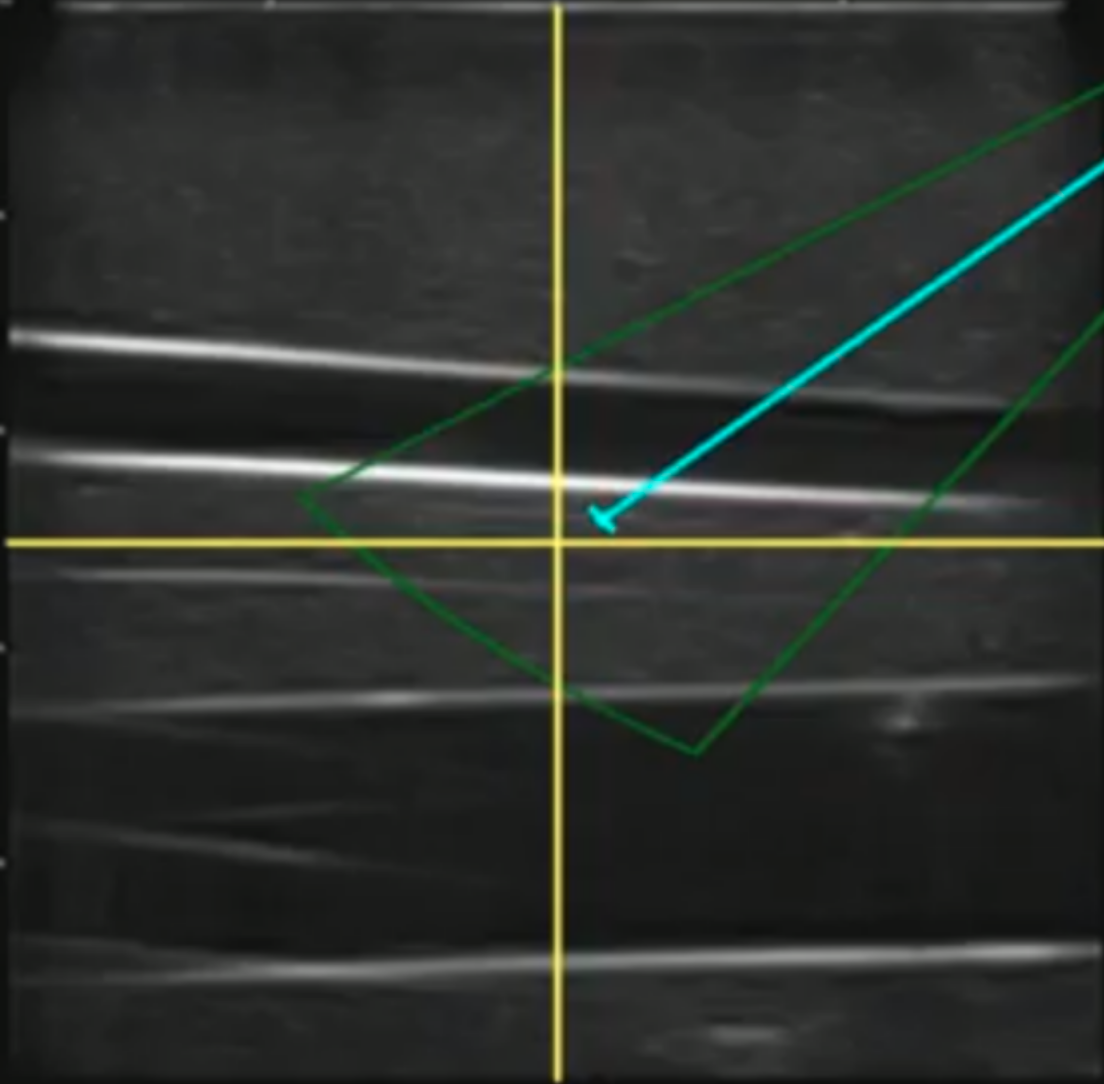}
        \caption{On the ultrasound image were visual overlays for click-based robot controller (yellow) and needle insertion controllers (cyan) and the needle insertion mechanism's workspace (green).}
        \label{fig:gui}
    \end{figure}
  
    Bridging the gap between the automated ultrasound scanning and needle insertion, an image-space controller was developed to augment the user's perception and control of the robot arm in conjunction with the ultrasound images. This system allowed for a single user to navigate the subdermal environment in order to view different, previously unobservable features. Previously we relied heavily on the UR3e pendant in conjunction with the monitor attached to the ultrasound, which overall lead to a divided operator attention and unintuitive control of the UR3e which required the user to command movement's in the robot's base frame. This process was necessary to align the ultrasound probe parallel to the femoral artery or vein as well as the surface of the skin as required for good ultrasound imaging and guide wire insertion as described by the Seldinger technique. 
    
    Our approach relied heavily on an augmented ultrasound image (Fig. \ref{fig:gui}). After a simple calibration which allowed us to convert ultrasound image space into real-world distance, a click on the image frame commanded the robot to move the ultrasound probe so that point was now located in the center of the frame. We also added additional buttons to control the rotation and non-planar position of the robot. This intuitive motion model made it possible for a user to simultaneously center the vessel in the center of the image and rotate until the vessel is transverse to the ultrasound probe (\ref{fig:gui}). 
    

\subsection{Needle Insertion (5-7)}

    The needle control algorithm is designed to take a user-selected point in the ultrasound image and drive the tip of the needle to that target. To aid the user in this task, we used the GUI to add an overlay of the work space of the needle, spanning the limits of the linear and angular degrees of freedom (Fig. \ref{fig:gui}). By clicking on the image within this workspace, the user could quickly and intuitively drive the needle anywhere within the work area. The calibration to calculate the transformation between actuator positions and ultrasound image space was done in a water bath.
    
    The transfer function between the pixel frame and the desired angular and linear position were derived from Fig. \ref{fig:ee_current} as follows:
        
        \begin{equation}
            x_{pixel} = (p_l l+l_{off})\cos(p_\theta\theta+\theta_{off})x_{scale} - x_{off}
            \label{eq:x_fk}
        \end{equation}
        \begin{equation}
            y_{pixel} =(p_l l+l_{off})\sin(p_\theta\theta+\theta_{off})y_{scale} - y_{off}
            \label{eq:y_fk}
        \end{equation}
        
    Where $l$ and $\theta$ are  the feedback from the angular and linear dofs, $p_l, l_{off}$, $p_\theta, \theta_{off}$ are parameters in linear functions to convert the actuator feedback to insert length and angle. $x_{off}$ and $y_{off}$ are the location of the center of rotation in pixel space and $x_{scale}$ and $y_{scale}$ are ultrasound scaling factors due to distortions of the ultrasound image. In total this gives us 8 parameters that were calculated by fitting to a calibration routine that sweep through possible linear and angular positions. More details about the needle tip calibration can be found in \cite{Robots:Wanwen2021}.
    
    In practice we found that this calibration was accurate enough for user control. There is a wealth of literature dedicated to the tracking of needle bending using ultrasound images \cite{needle_stearing1, needle_stearing2, Robots:Wanwen2021}. Although due to complex needle-tissue interactions this kind of complex needle tracking will be necessary for full automation of needle insertion, it was accurate enough for the operator to use as a visual aid when planning a needle insertion, Fig. \ref{fig:gui}. 
    
    To correct errors in the kinematic estimate of the needle tip the users also relied on a so called "needle tweak controller." By clicking along the axis of the needle the user could adjust the needle depth in the ultrasound frame by visually estimating the error in the kinematics.
    
\section{Results}

    To test our system, we performed experiments on both a silicone tissue phantom and in-vivo porcine surgeries. These experiments are not a substitute for a true user study and are not meant to generalize to the general population, as the operators were each intimately familiar with at least one part of the system and are authors of this paper. Rather, the goal of these experiments is primarily to show the viability of our approach and the efficacy of our system. In both kinds of experiments, placement of a guidewire was our standard for success. The reason for this was that,  unlike systems for drawing blood, the needle must be placed very close to the center of the vessel as well as at a shallow enough angle for the guidewire and later a catheter to be placed correctly \cite{Seldinger:Seldinger1953}. A cannulation near the side of the vessel wall or at too steep an angle would result in the guidewire getting caught on or even puncturing the vessel wall. 

\subsection {Tissue phantom experiments} \label{phantom_results}

        \begin{table}[htbp]
            \caption{Phantom Insertion Time Trials}\label{tab:kinematic_calib_error}
            
            \begin{center}
                \begin{tabular}{c|c|c|c}
                & \textbf{Trial 1} & \textbf{Trial 2} & \textbf{Trial 3} \\
                \hline
                \textbf{User 1} & 4m 06s & 5m 05s& 4m 01s \\
                \hline
                \textbf{User 2} & 3m 49s & 3m 30s & 3m 59s\\ 
                \hline
                \textbf{User 3} & 7m 39s & 5m 45s & 5m 58s\\ 

            \end{tabular}
            \label{tab:results}
            \end{center}
        \end{table}

    We tested our system on a Blue Phantom Generation II Femoral lower torso training model. This allowed us to test the GUI, scanning and needle insertion in a more controlled environment than that of in-vivo studies, as well as to study the efficacy of our system on human anatomy.
    
    These experiments were designed to test the system's repeatably and show its efficacy when used by different operators. Three operators were first trained to perform the Seldinger technique by hand to ensure that they each had a basic understanding of the procedure. Then, each operator was given a basic explanation of the GUI and control scheme. Each operator then followed the process described in section \ref{procedure} and were given three attempts to insert the needle into the artery and place a guide wire. An attempt was considered a success when the guide wire was fed all the way to the end the phantom's artery and pulled clear of the mechanism. With these instructions each user was able to complete the procedures on a phantom without failing and in times summarized in Tb. \ref{tab:results}.
    

\subsection{In-vivo results}

    \begin{figure}[thpb]
        \centering
        \includegraphics[width=3in]{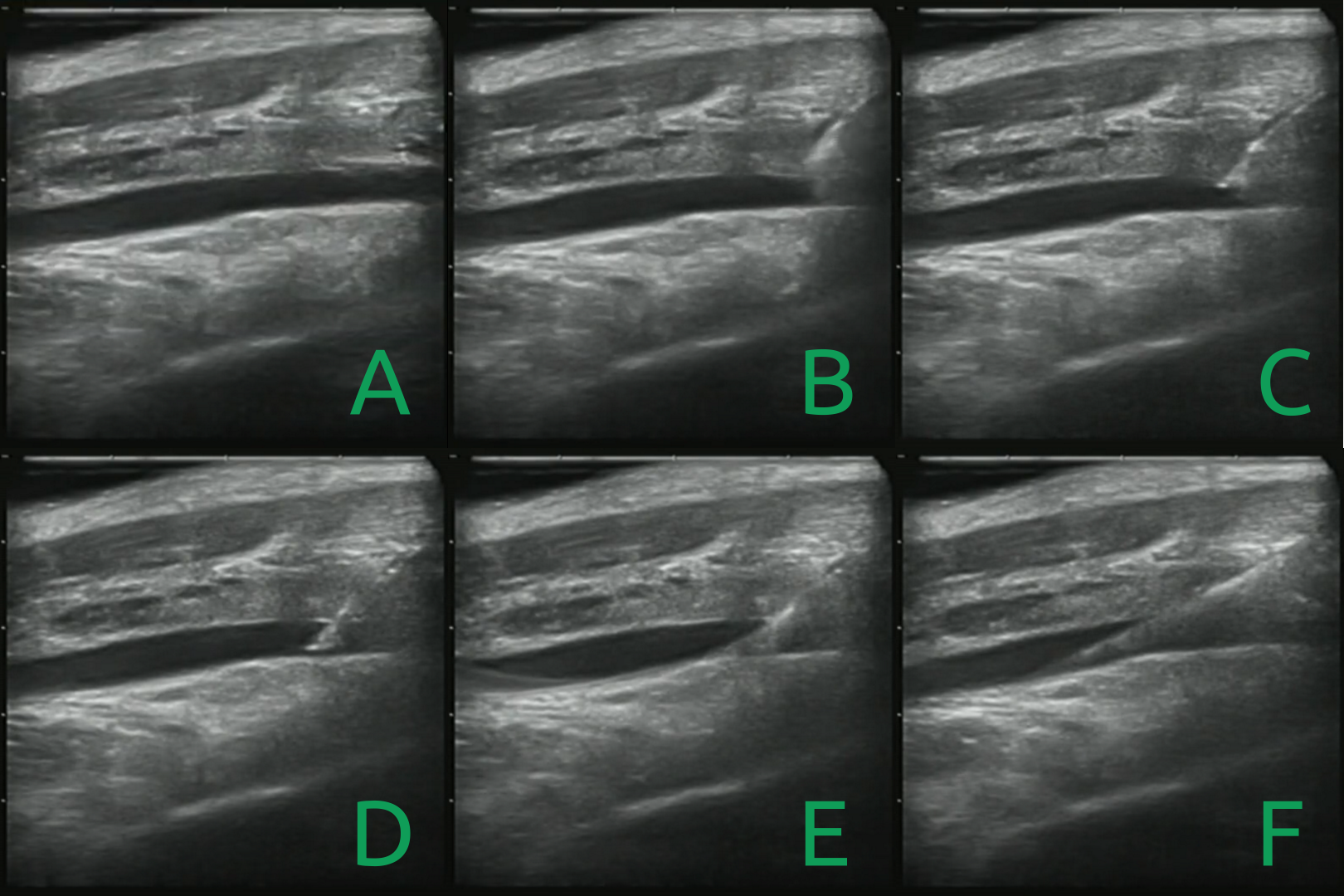}
        \caption{Ultrasound images captured during successful guidewire insertion. A shows the target vein before intervention. C shows the needle pushing on the vein wall. D shows the needle clearly inside of the lumen. E shows guidewire insertion through the needle. F shows guidewire placement after needle has been retracted and removed.}
        \label{fig:success}
    \end{figure}
    
    This system presented in this paper was also used in three in-vivo porcine surgeries. In each experiment, an operator followed the procedure described in section \ref{procedure}. In addition, once the guidewire was placed, a catheter was slid over the guidewire and the guidewire was removed by the operator, completing the full procedure for femoral access.

    These experiments differed from those performed on the phantom in several ways. First, the vasculature was much more prone to rolling,  and deforming under the needle than in the phantom. In addition, the non-uniformity of the tissue could cause the needle to bend or deflect slightly. Finally, the geometry of the femoral triangle is much more curved near the femoral triangle in pigs than it is in humans.
    
    Despite these challenges, the operator was able to use the system cannulate the femoral artery, place a guidewire, and use that guidewire to place a catheter in each of the three pigs, without causing complications such as lacerations or hematomas. While this may not be a particularly impressive feat for a trained surgeon, we consider it a great success considering the operators had no formal medical training. Though more experimentation must be undertaken to fully assess the efficacy and usability of the system, these experiments show the system's real world operablility, and is a significant step in our goal of fully automating this procedure.

\section{CONCLUSION}

    In this paper we present a system capable of teleoperated needle insertion. This needle insertion allowed for subsequent guide wire and catheter insertion. The main contributions of this paper are the development of a robotic system capable of assisting a human with this medical procedure in a way that is reliable enough to test in-vivo. By understanding how a human will interact with this device we can now continue to collect data at each state of the insertion process and ultimately decide to either further assist or phase out the human operator. Moving forward, our biggest challenges will likely be in the realm of sensing and computer vision; generating enough labeled sub-dermal images and needle insertion data for the development of models capable of understanding ultrasound, puncture forces, and the dynamic environment of subdermal anatomy. Currently, the most time consuming part of the procedure is aligning the robot with the artery in the longitudinal plane, especially on pigs where the vasculature is more curved than on humans. In order to automate this process, we will need to reconstruct a 3D model of the vessels and surface of the skin. In addition, while a flush of blood indicating a successful cannulation is guaranteed in the femoral artery of a healthy patient, patients with low blood pressure such as those in trauma will require more sensitive methods to detect puncture such as force, bio-impedance, or pressure sensors.

\section*{ACKNOWLEDGMENTS}

	Thank you first and foremost to the pigs. 
    A huge thank you to Lisa Gordon, Antonio Gumucio, Ted Lagattuta for all of their help. Furthermore this work would not be possible without Nate Shoemaker-Trejo and Charlie Hart who contributed to the design and manufacturing of the mechanism. Finally we would like to thank Tejas Zodage for his help with ultrasound calibration. 
 
\bibliographystyle{bib/IEEEtran} 
\bibliography{bib/bibliography}

\end{document}